\documentclass[conference]{IEEEtran}
\IEEEoverridecommandlockouts

\usepackage{cite}
\usepackage{amsmath,amssymb,amsfonts}
\usepackage{algorithmic}
\usepackage{graphicx}
\usepackage{textcomp}
\usepackage{xcolor}

\usepackage{float}
\usepackage{graphicx}
\usepackage{tabularx}
\usepackage{multirow}
\usepackage{multicol}
\usepackage{booktabs}
\usepackage{amssymb}
\usepackage{pifont}
\usepackage{subcaption}
\usepackage{bm}
\usepackage[english]{babel}
\usepackage{microtype}
\usepackage{enumitem}
\usepackage{subcaption}
\usepackage{titlesec}
\usepackage{color}
\usepackage{array}
\usepackage[hidelinks]{hyperref}

\def\BibTeX{{\rm B\kern-.05em{\sc i\kern-.025em b}\kern-.08em
    T\kern-.1667em\lower.7ex\hbox{E}\kern-.125emX}}
\begin{document}

\title{
    Diffusion Features to Bridge Domain Gap for Semantic Segmentation
}

\author{\IEEEauthorblockN{Yuxiang Ji$^\star$}
\IEEEauthorblockA{\textit{Institute of Artifcial Intelligence} \\
\textit{Xiamen University}\\
Xiamen, China \\
yuxiangji@stu.xmu.edu.cn}
\and
\IEEEauthorblockN{Boyong He$^\star$\thanks{$\star$ Contribute equally to the work.}}
\IEEEauthorblockA{\textit{Institute of Artifcial Intelligence} \\
\textit{Xiamen University}\\
Xiamen, China \\
boyonghe@stu.xmu.edu.cn}
\and
\IEEEauthorblockN{Chenyuan Qu}
\IEEEauthorblockA{\textit{School of Computer Science} \\
\textit{University of Birmingham}\\
Birmingham, United Kingdom \\
cxq134@student.bham.ac.uk}
\and
\IEEEauthorblockN{Zhuoyue Tan}
\IEEEauthorblockA{\textit{Institute of Artifcial Intelligence} \\
\textit{Xiamen University}\\
Xiamen, China \\
tanzhuoyue@stu.xmu.edu.cn}
\and
\IEEEauthorblockN{Chuan Qin}
\IEEEauthorblockA{\textit{Department of Infrastructure Engineering} \\
\textit{The University of Melbourne}\\
Melbourne, Australia \\
qincq1@student.unimelb.edu.au}
\and
\IEEEauthorblockN{Liaoni Wu$^\dag$\thanks{$\dag$ Corresponding author.}}
\IEEEauthorblockA{\textit{Institute of Artifcial Intelligence} \\
\textit{Xiamen University}\\
Xiamen, China \\
wuliaoni@xmu.edu.cn}
}

\maketitle

\begin{abstract}
    Pre-trained diffusion models have demonstrated remarkable proficiency in synthesizing images across a wide range of scenarios with customizable prompts, indicating their effective capacity to capture universal features. 
    Motivated by this, our study delves into the utilization of the implicit knowledge embedded within diffusion models to address challenges in cross-domain semantic segmentation. 
    This paper investigates the approach that leverages the sampling and fusion techniques to harness the features of diffusion models efficiently. 
    We propose DIffusion Feature Fusion (DIFF) as a backbone use for extracting and integrating effective semantic representations through the diffusion process.
    By leveraging the strength of text-to-image generation capability, we introduce a new training framework designed to implicitly learn posterior knowledge from it.
    Through rigorous evaluation in the contexts of domain generalization semantic segmentation, we establish that our methodology surpasses preceding approaches in mitigating discrepancies across distinct domains and attains the state-of-the-art (SOTA) benchmark.
    The implementation code is released at \url{https://github.com/Yux1angJi/DIFF}.
\end{abstract}
\vspace{-0.3cm}
\section{Introduction}

The paradigm of training segmentation models on large-scale datasets has demonstrated significant successes; nevertheless, the obstacles associated with the acquisition of data specific to niche scenarios, continue to pose significant challenges.
Synthetic data, while complementing some missing data, usually suffers from the issue of domain gaps.
This issue arises because models trained on limited synthetic data tend to decline in accuracy when applied in real-world settings, attributable to domain shifts in the test data~\cite{wangGeneralizingUnseenDomains2022, zhouDomainGeneralizationSurvey2022}.
Research has shown that one of the important factors is the representation discrepancy caused when the perspective, background, style, or imaging conditions are changed to the unseen domain~\cite{namReducingDomainGap2021, gaoAddressingDomainGap2021, ganinUnsupervisedDomainAdaptation2015}.
Taking this issue, the study of Domain Generalization (DG) focuses on reducing the domain variance performance and improving the model robustness across unseen domains.

Recently, the stunning performance of diffusion models on various tasks of image generation has attracted a great deal of research attention. 
The pre-trained text-to-image diffusion model (e.g., Stable Diffusion~\cite{rombachHighResolutionImageSynthesis2022}) possesses the capability to synthesis images of remarkable realism and high quality across diverse styles, scenes, and categories, contingent upon the customized prompts given. 
This indicates that the diffusion model learns generic visual features while being able to disentangle the representations of image features according to conditional text inputs. 
Several studies also validate the ability of pre-trained diffusion models on representational and perceptual tasks~\cite{zhaoUnleashingTexttoImageDiffusion2023, baranchukLabelEfficientSemanticSegmentation2022, mukhopadhyayDiffusionModelsBeat2023, chenDeconstructingDenoisingDiffusion2024, luoDiffusionHyperfeaturesSearching2023, tangEmergentCorrespondenceImage2023}. 
Drawing inspiration from the implicit universal knowledge embedded within pre-trained diffusion models, it leads us to think: \textit{how to utilize such knowledge to reduce the domain discrepancy in semantic segmentation?}

We therefore introduce \textbf{DI}ffusion \textbf{F}eature \textbf{F}usion (DIFF), a module based on the pre-trained Stable Diffusion model, to collect and integrate the feature sets from the whole diffusion process.
Given an image from arbitrary domains, robust features modeled in the diffusion embedding space could be extracted from the Stable Diffusion.
Different from similar studies~\cite{gongPromptingDiffusionRepresentations2023,xuOpenVocabularyPanopticSegmentation2023} that use single-step denoising for extracting features, we consider the diffusion trajectory as a more meaningful feature.
The feature sets from the whole multi-step diffusion process will be fused by a convolutional fusion block and aligned to the visual embedding from the standard backbones, e.g. ResNet~\cite{heDeepResidualLearning2016}, Vision Transformer~\cite{dosovitskiy2020image}.
To further utilize the conditional generation capability of the pre-trained diffusion model and to address the absence of corresponding annotation text as conditional input when doing prediction, we introduce a special implicit posterior knowledge learning framework for supervised learning. 
By leveraging vision-language joint modeling, the implicit posterior knowledge from the conditional generation capability of diffusion models could be learned to maintain generalization when confronted with unseen data.

\begin{figure*}[!ht]
    \centering
    \includegraphics[width=\textwidth]{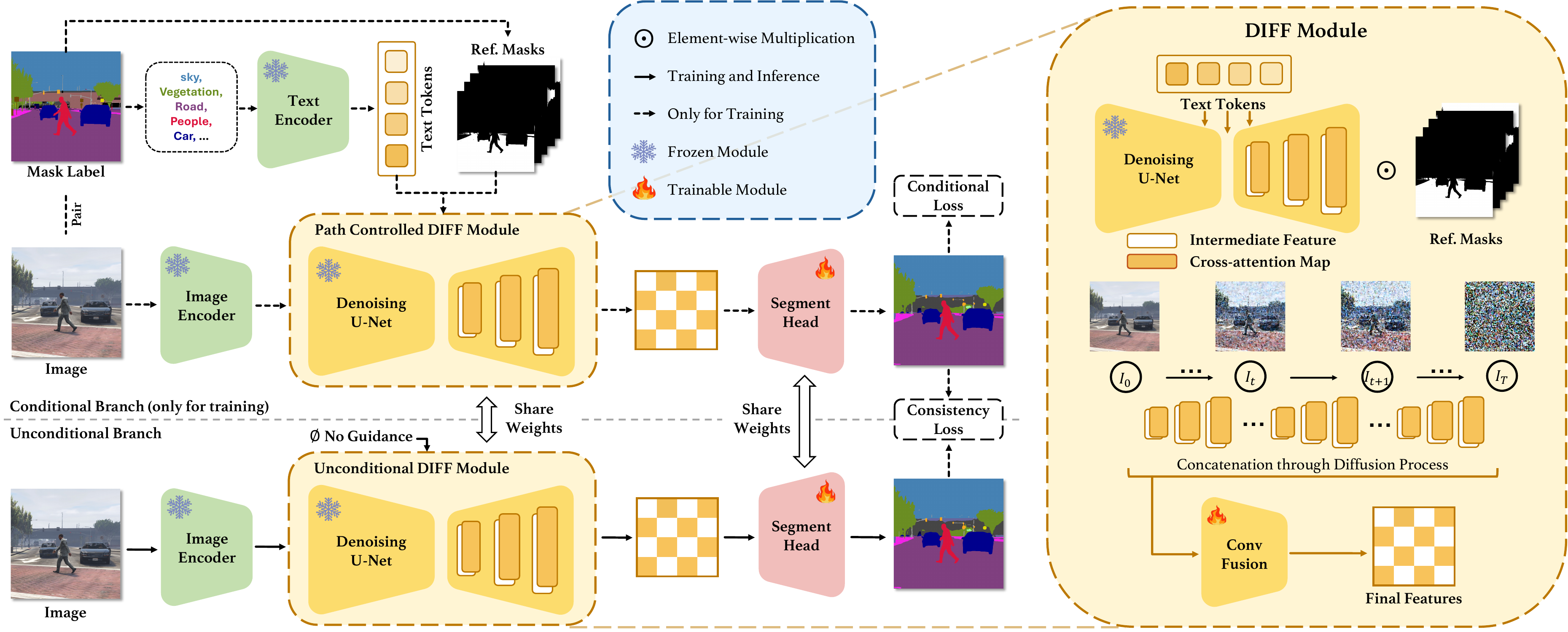} 
    \caption{
        Overview of proposed diffusion feature fusion (DIFF) module and implicit posterior knowledge learning (IPKL) training pipeline. 
        (a) In the conditional branch, we extract the categories and reference masks from the semantic segmentation annotations and use them as conditions, which are input into the DIFF module along with the input image. 
        Through the DIFF module, we obtain features enhanced with conditional information for supervised training.
        (b) In the unconditional branch, we only use the image as input to the DIFF module and employ the prediction results from the conditional branch as a teacher for consistency learning.
    }
    \label{fig:pipeline}
\end{figure*}

\section{Method}
By given a text-image paired dataset, the original generation-purpose text-to-image diffusion model is set up to model a process that gradually removes noise from a standard noise distribution to an image distribution based on text prompt, or in other words, it possesses the conditional distribution $p(\mathcal{X}|\mathcal{C})$.
As for standard segmentation tasks, let $\{\mathcal{X,Y}\}$ be the set of images and mask labels, the target is to learn $p(\mathcal{Y}|\mathcal{X})$.
Considering text as a higher-level understanding, it is straight-froward to leverage the prior knowledge $p(\mathcal{X}|\mathcal{C})$ of pre-trained text-to-image diffusion models to help learn a more generalized posterior distribution for segmentation models.

Therefore, like some recent perceptual works based on diffusion models~\cite{tangEmergentCorrespondenceImage2023,luoDiffusionHyperfeaturesSearching2023,zhaoUnleashingTexttoImageDiffusion2023,xuOpenVocabularyPanopticSegmentation2023,baranchukLabelEfficientSemanticSegmentation2022,gongPromptingDiffusionRepresentations2023}, we rely on the critical denoising component U-Net~\cite{ronneberger2015u} for a more generalized encoding.
Unlike the usual generative tasks where the noise predicted by U-Net is gradually removed, we perform an inverse process, that the predicted noise is gradually added to the input image.
Different from similar studies~\cite{gongPromptingDiffusionRepresentations2023,zhaoUnleashingTexttoImageDiffusion2023} that use single-step denoising for extracting features, we consider the whole multi-step diffusion trajectory as a more meaningful feature.
In order to retain more information from pre-trained diffusion model without disruption and to ensure that the features can be utilized by the segmentation head model, we freeze the entire U-Net and sample as many effective features as possible for fusion, as detailed in Sec.~\ref{sec:diff}.
On the other hand, considering that standard segmentation tasks do not have text prompts as input, we introduce a dual-stream network for learning posterior knowledge to perform conventional segmentation predictions, as detailed in Sec.~\ref{sec:ipkl}.
The overview of the proposed pipeline is shown in Fig.~\ref{fig:pipeline}, where the training is performed on two branches and the prediction is only performed on the unconditional branch.
The trainable modules of the two branches share weights.

\subsection{Diffusion Feature Fusion}
\label{sec:diff}
In alignment with the architectural design and pre-training scheme of diffusion model, we devise two distinct feature sets from U-Net aimed at facilitating the integration of visual and text semantic understanding.
Specifically, we extract the intermediate features $\{\mathcal{F}_{t, l}^\text{inter}\} \in \mathbb{R}^{d_l \times w_l \times h_l}$ from each layer $l$ and each step $t$ within the U-Net decoder as the visual representation, and the cross-attention maps $\{\mathcal{F}_{t, l}^\text{cross}\} \in \mathbb{R}^{d_l \times w_l \times h_l}$ as the interaction representations between visual and text content. 
However, directly taking out the features of the whole process layer by layer would lead to too numerous and unwieldy for practical utilization, which is also the reason that many works apply a one-step diffusion pipeline and hand-select the block, step pairs by grid search.
Inspired by DiffHyperfeature~\cite{luoDiffusionHyperfeaturesSearching2023}, here we propose DIffusion Feature Fusion (DIFF) to fuse features extracted from different layers with different steps by an aggregating convolutional block, including a series of convolution, normalization, and activation layers.

Formally, given intermediate features $\{\mathcal{F}_{t,l}^\text{inter}\}$ and cross-attention maps $\{\mathcal{F}_{t,l}^\text{cross}\}$ from the diffusion process of the input images, the final features for segmenting $\mathcal{F}_\text{diff}$ takes the form
\begin{equation}
    \mathcal{F}_\text{diff} = F_\text{conv}(\oplus_{t,l} [\mathcal{F}_{t, l}^\text{inter}, \mathcal{F}_{t, l}^\text{cross}]),
\end{equation}
where $\oplus_{t,l}$ means the concatenation across $t$ and $l$.

\begin{figure*}[!t]
    \centering
    \includegraphics[width=\textwidth]{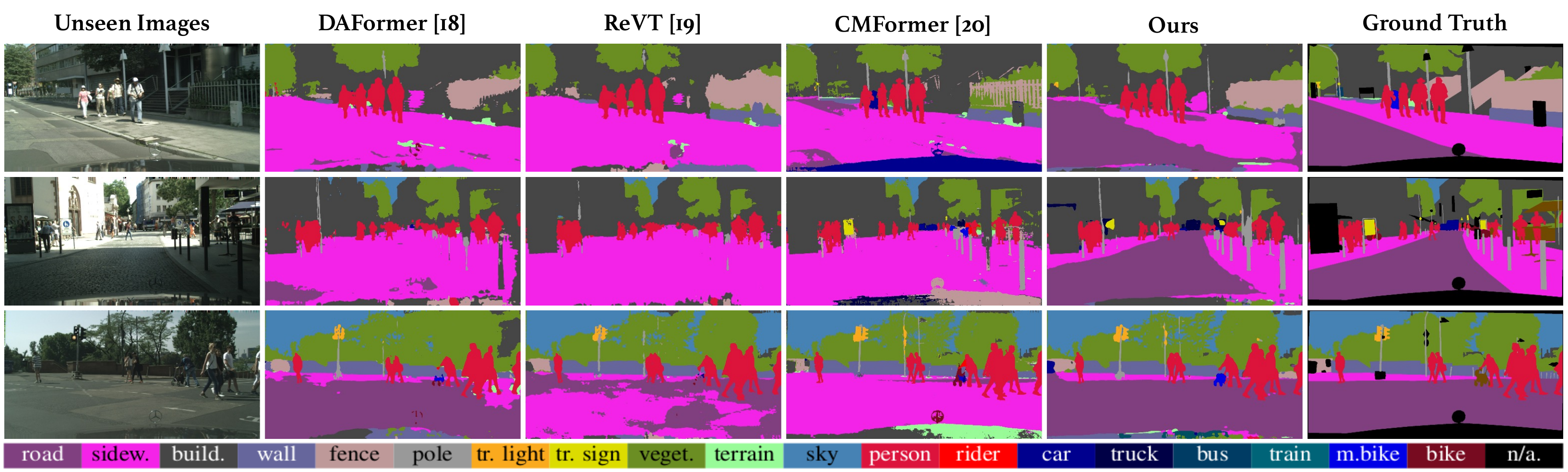} 
    \caption{
        Segmenting prediction on the unseen data of existing SOTA domain generalization (DG) semantic segmentation methods (DAFormer\cite{hoyerDAFormerImprovingNetwork2022}, ReVT\cite{termohlenReParameterizedVisionTransformer2023}, CMFormer\cite{biLearningContentenhancedMask2023a}) and our method.
    }
    \label{fig:seg cmp}
\end{figure*}

\subsection{Implicit Posterior Knowledge Learning}
\label{sec:ipkl}
In order to learn posterior knowledge $p(\mathcal{Y}|\mathcal{X})$ for standard semantic segmentation from the prior knowledge $p(\mathcal{X}|\mathcal{C})$ of pre-trained diffusion models, we still have two unresolved issues:
(1) The first issue is that original text-to-image diffusion models are set up to model the relationship between the overall textual description and the image, which leaves a gap to the pixel-level fine-grained segmentation.
(2) The second issue is that there are no reference guided input provided when predicting in the standard semantic segmentation tasks.
To address these two issues, we adopt path controlled diffusion in Sec.~\ref{sec:path controll} and unconditional consistency learning in Sec.~\ref{sec:consistency}.

\subsubsection{Path Controlled Diffusion}
\label{sec:path controll}
To ensure the extracted features correspond more precisely with the textual prompts, we adopt a training-free method following MultiDiffusion~\cite{bar-talMultiDiffusionFusingDiffusion2023}.
By given segmentation data $\{\mathcal{X,Y}\}$ during training, where $\mathcal{X}$ are images and $\mathcal{Y}$ are pixel-level semantic annotations, we decompose the annotations $\mathcal{Y}$ into multiple groups of masks $\mathcal{M} \in \{0,1\}^{\text{cls} \times w \times h}$ and their corresponding category descriptions $\mathcal{C} \in \{\textit{sky, vegetation, road, people, car, ...}\}^\text{cls}$, where cls represents the number of all categories.
Then we feed the category descriptions $\mathcal{C}$ as text prompts into the diffusion model and fuse the predicted latent variables by the corresponding masks $\mathcal{M}$ at each diffusion step, formulate as
\begin{equation}
    I_{t+1} (I_{t}, \mathcal{M,C}) = \sum_{i=1}^{\text{cls}}\frac{\mathcal{M}_i}{\sum_{j=1}^{\text{cls}} \mathcal{M}_j} [\mathcal{M}_i \odot (\Phi (I_{t}, \mathcal{C}_i))],
\end{equation}
where $\Phi$ represents the pre-trained text-to-image diffusion model and $I$ represent the latent variables.

Then by feeding such conditional features extracted by DIFF to the segment head $D$, we could train the entire model using cross-entropy, just like in conventional segmentation models as
\begin{equation}
    \mathcal{L}_\text{condit} = \text{CE}(D(\mathcal{F}_\text{diff}^\text{con}(\mathcal{X,M,C})), \mathcal{Y}).
\end{equation}

\subsubsection{Unconditional Consistency Learning}
\label{sec:consistency}
To address the issue of not having corresponding text prompts during prediction, we set up an additional unconditional branch as shown in Fig.~\ref{fig:pipeline}.
In this branch, the diffusion features $\mathcal{F}_\text{diff}^\text{uncon}$ are acquired with DIFF through an unconditional process $p(\mathcal{F} | \mathcal{X}, \mathcal{C}=\emptyset)$.
On this basis, we simply employ a L2 loss for consistency learning between the outputs from two branches, forcing them to produce nearly identical outputs, as
\begin{equation}
    \mathcal{L}_\text{consis} = \| D(\mathcal{F}_\text{diff}^\text{con}(\mathcal{X,M,C})), D({\mathcal{F}_\text{diff}^\text{uncon}}(\mathcal{X})) \|_2.
\end{equation}
In this way, the implicit posterior knowledge learned from the conditional generative model could be gradually distilled onto the unconditional input branch for predicting.
The complete learning objective is the combination of conditional segmentation loss $\mathcal{L}_\text{condit}$ and consistency loss $\mathcal{L}_\text{consis}$ as
\begin{equation}
    \mathcal{L}_\text{final} = \lambda_1 \mathcal{L}_\text{condit} + \lambda_2 \mathcal{L}_\text{consis}.
\end{equation}
Here, $\lambda_1$ and $\lambda_2$ are two hyper-parameters to control the weights of two objectives, which are both set to 1 in our experiments.

\section{Experiments}

\subsection{Experimental Setup}
Following the setup of prior research on domain generalization for semantic segmentation~\cite{hoyerDAFormerImprovingNetwork2022,huangFSDRFrequencySpace2021,gongPromptingDiffusionRepresentations2023}, the model trained on a source domain dataset will be evaluated on a series of unknown target datasets.
Based on two synthetic datasets GTAV~\cite{richterPlayingDataGround2016} and Synthia~\cite{rosSYNTHIADatasetLarge2016}, three real-world clear datasets Cityscapes (CS)~\cite{cordtsCityscapesDatasetSemantic}, BDD-100K (BDD)~\cite{yuBDD100KDiverseDriving2020}, and Mapillary Vistas (MV)~\cite{neuholdMapillaryVistasDataset}, and two real-world adverse weather datasets ACDC~\cite{sakaridisACDCAdverseConditions2021} and Dark Zurich (DZ)~\cite{sakaridisGuidedCurriculumModel2019}, we consider two practically significant generalization tasks: \textit{synthetic-to-real} and \textit{real-to-adverse}.
\textit{Synthetic-to-real}: We use two synthetic datasets GTAV and Synthia as source domain, and five real-world datasets CS, BDD, MV, ACDC, and DZ as target domain.
\textit{Real-to-Adverse}: We use clear dataset CS as source domain, and two adverse weather datasets ACDC and DZ as target domain.

For implementation, our DIFF module is based on the released Stable-Diffusion v2-1~\cite{rombachHighResolutionImageSynthesis2022} checkpoint, and we use the same decoder head and training settings as in DAFormer~\cite{hoyerDAFormerImprovingNetwork2022}.

\paragraph*{Remark:} In tables, the best results are highlighted in \textbf{bold}, while the second best is \underline{underlined}.

\subsection{Results}
\begin{table}[t]
    \centering
    \caption{Synthetic-to-real DG results comparing to SOTA methods on GTAV source domain.
        Training is performed on synthetic dataset GTAV~\cite{richterPlayingDataGround2016}.
        Evaluation is performed on five real-world datasets with 19 categories.
    }
    \label{tbl:main result gta}
    \resizebox{\columnwidth}{!}{
      \begin{tabular}{lc|ccc|c|cc|c}
          \toprule
          \multirow{2}{*}{DG Method}  &   \multirow{2}{*}{Backbone} &  \multicolumn{7}{c}{mIoU (\%) on} \\
          \cmidrule{3-9}
                                      &               &   CS~\cite{cordtsCityscapesDatasetSemantic}   &     BDD~\cite{yuBDD100KDiverseDriving2020}      &   MV~\cite{neuholdMapillaryVistasDataset}    &   Avg3                 &   ACDC~\cite{sakaridisACDCAdverseConditions2021}                &   DZ~\cite{sakaridisGuidedCurriculumModel2019}    & Avg5    \\
          \midrule
          IBN-Net~\cite{panTwoOnceEnhancing2020}                       &   ResNet-101  &  37.37              &  32.29              &  33.84              &   33.15                &    -                  &    -    &   -     \\
          FSDR~\cite{huangFSDRFrequencySpace2021}                      &   ResNet-101  &  44.80              &  41.20              &  43.40              &   43.13                &   24.77               &   9.66  &  32.77  \\
          SAN-SAW~\cite{pengSemanticAwareDomainGeneralized2022}        &   ResNet-101  &  45.33              &  41.18              &  40.77              &   42.23                &    -                  &    -    &   -     \\
          ReVT~\cite{termohlenReParameterizedVisionTransformer2023}    &   MiT-B5      &  49.96              &  48.01              &  53.06              &   50.34                &   41.15               &   21.99 &  42.84  \\
          DAFormer~\cite{hoyerDAFormerImprovingNetwork2022}            &   MiT-B5      &  52.65              &  47.89              &  54.66              &   51.73                &   38.25               &   17.45 &  42.18  \\
          CMFormer~\cite{biLearningContentenhancedMask2023a}           &   SWin-B      &  \underline{55.31}  &  \underline{49.91}  &  \textbf{60.09}     &   \underline{55.10}    &   \underline{41.34}   &   \underline{22.58} &  \underline{45.85}  \\
          PromptDiff~\cite{gongPromptingDiffusionRepresentations2023}  &   Diffusion   &  52.00              &    -                &    -                &     -                  &    -                  &    -    &   -     \\
          \midrule
          Ours                                                         &   Diffusion   &  \textbf{58.01}     &  \textbf{53.60}     &  \underline{59.85}  &   \textbf{57.15}       &   \textbf{46.32}      &   \textbf{34.27} &  \textbf{50.41}  \\
                                                                       &               &  \textcolor{red}{+2.70}  &  \textcolor{red}{+3.69}  &  \textcolor{green}{-0.24}  &  \textcolor{red}{+2.05}  &  \textcolor{red}{+4.98}  &  \textcolor{red}{+11.69}  &  \textcolor{red}{+4.56}  \\
          \bottomrule
      \end{tabular}
    }
\end{table}

\begin{table}[t]
    \centering
    \caption{
        Synthetic-to-real DG results comparing to SOTA methods on Synthia source domain.
        Training is performed on synthetic dataset Synthia~\cite{rosSYNTHIADatasetLarge2016}.
        Evaluation is performed on five real-world datasets with 16 categories.
    }
    \label{tbl:main result synthia}
    \resizebox{\linewidth}{!}{
    \begin{tabular}{lc|ccc|c|cc|c}
        \toprule
        \multirow{2}{*}{DG Method}  &   \multirow{2}{*}{Backbone}    &  \multicolumn{7}{c}{mIoU (\%) on} \\
        \cmidrule{3-9}
        &               &   CS~\cite{cordtsCityscapesDatasetSemantic}   &     BDD~\cite{yuBDD100KDiverseDriving2020}      &   MV~\cite{neuholdMapillaryVistasDataset}    &   Avg3                 &   ACDC~\cite{sakaridisACDCAdverseConditions2021}                &   DZ~\cite{sakaridisGuidedCurriculumModel2019}    & Avg5    \\
        \midrule
        IBN-Net~\cite{panTwoOnceEnhancing2020}                        &   ResNet-101  &  32.04  &  30.57  &  32.16  &   31.59    &    -      &    -    &   -     \\
        FSDR~\cite{huangFSDRFrequencySpace2021}                       &   ResNet-101  &  40.80  &  39.60  &  37.40  &   39.30    &    -      &    -    &   -  \\
        SAN-SAW~\cite{pengSemanticAwareDomainGeneralized2022}         &   ResNet-101  &  38.92  &  35.24  &  34.52  &   36.23    &    -      &    -    &   -     \\
        DAFormer~\cite{hoyerDAFormerImprovingNetwork2022}             &   MiT-B5      &  44.08  &  33.20  &  42.99  &   40.09    &   26.62   &   14.14  &  32.21  \\
        ReVT~\cite{termohlenReParameterizedVisionTransformer2023}     &   MiT-B5      &  46.28  &  \underline{40.30}  &  \underline{44.76}  &   \underline{43.78}    &    \underline{35.75}      &    \underline{20.10}    &   \underline{37.44}  \\
        CMFormer~\cite{biLearningContentenhancedMask2023a}            &   SWin-B      &  44.59  &  33.44  &  43.25  &   40.43   &   34.50   &  19.57  &   35.07   \\
        PromptDiff~\cite{gongPromptingDiffusionRepresentations2023}   &   Diffusion   &  \underline{49.10}  &    -    &    -    &     -      &     -     &    -     &   -     \\   
        \midrule
        Ours                                                          &   Diffusion   &  \textbf{49.31}  &  \textbf{42.20}  &  \textbf{49.47}  &    \textbf{46.99}   &   \textbf{36.27}   &  \textbf{23.39}  &  \textbf{40.13}  \\
                                                                        &               &  \textcolor{red}{+0.21}  &  \textcolor{red}{+1.90}  &  \textcolor{red}{+4.71}  &  \textcolor{red}{+3.21}  &  \textcolor{red}{+0.52}  &  \textcolor{red}{+3.29}  &  \textcolor{red}{+2.69}  \\
        \bottomrule
    \end{tabular}
    }
\end{table}

Under the \textit{synthetic-to-real} setting as shown in Tab.~\ref{tbl:main result gta}, our method is demonstrated to outperform the previous SOTA DG methods based on ResNet-101~\cite{heDeepResidualLearning2016}, MiT-B5~\cite{xieSegFormerSimpleEfficient2021}, SWin-B~\cite{liu2021swin}, and diffusion backbones.
Specifically, our method shows higher improvements (4.98\% and 11.69\%) on the two adverse weather datasets ACDC and DZ, that the data with larger domain discrepancies.
Comparing to the PromptDiff~\cite{gongPromptingDiffusionRepresentations2023} which is also a diffusion-based method, our performance surpasses it by 6.01\% on GTAV $\rightarrow$ CS.  
We also extend the experiments on the synthia source domain in Tab.~\ref{tbl:main result synthia} for further validation.

Since our method freezes most of the network parameters, the learning capacity in the source domain could be limited.
Under the \textit{real-to-adverse} setting as shown in Tab.~\ref{tbl:main result cs}, in settings with smaller domain differences (i.e., CS $\rightarrow$ ACDC), our method is inferior (-1.30\%) to those based on a transformer backbone, but it shows significant improvement (+4.38\%) in settings with larger domain differences (i.e., CS $\rightarrow$ DZ).

Fig.~\ref{fig:seg cmp} provides some visual examples of semantic segmentation results from our method comparing to the previous SOTA DG methods.
It shows that our method improves the generalization on the unseen data to a large extent, especially on the \textit{road, sidewalk} categories.

\begin{table}[t]
    \centering
    \caption{
        Clear-to-adverse DG results comparing to SOTA methods on Cityscapes source domain.
        Training is performed on clear dataset Cityscapes\cite{cordtsCityscapesDatasetSemantic}.
        Evaluation is performed on two adverse weather datasets with 19 categories.
    }
    \label{tbl:main result cs}
    \resizebox{0.8\linewidth}{!}{
    \begin{tabular}{lc|cc|c}
        \toprule
        \multirow{2}{*}{DG Method}  &   \multirow{2}{*}{Backbone}    &  \multicolumn{3}{c}{mIoU (\%) on} \\
        \cmidrule{3-5}
        &           &   ACDC~\cite{sakaridisACDCAdverseConditions2021}     &   DZ~\cite{sakaridisGuidedCurriculumModel2019}    &  Avg2    \\
        \midrule
        FSDR~\cite{huangFSDRFrequencySpace2021}  & ResNet-101 &   47.18    &   22.60  &  34.89  \\
        SAN-SAW~\cite{pengSemanticAwareDomainGeneralized2022}  & ResNet-101 &   49.00    &   24.80  &  36.90  \\
        DAFormer~\cite{hoyerDAFormerImprovingNetwork2022}   & MiT-B5   &   55.15    &   28.28  &  41.72  \\
        CMFormer~\cite{biLearningContentenhancedMask2023a}  & SWin-B       &   \textbf{60.21}    &   33.90  &  \underline{47.05}  \\
        PromptDiff~\cite{gongPromptingDiffusionRepresentations2023}  & Diffusion     &   58.60    &   \underline{34.00}  &  46.30  \\
        \midrule
        Ours   & Diffusion         &   \underline{58.91}    &   \textbf{38.38}  &  \textbf{48.65}  \\
                                   &               &  \textcolor{green}{-1.30}  &  \textcolor{red}{+4.38}  &  \textcolor{red}{+1.60}  \\
        \bottomrule
    \end{tabular}
    }

\end{table}

\subsection{Ablation Study}
We evaluate the effects of proposed diffusion feature fusion (DIFF) and implicit posterior knowledge learning (IPKL) with different consistency losses on the overall results in Tab.~\ref{tbl:ablation study}.
The results indicate that both DIFF and IPKL could improve the domain generalization performance of the model, while only introducing the IPKL training framework without using consistency loss leads to significant performance degradation.
As for the choice of consistency loss function, both L2 loss and KL loss contribute to improving consistency learning, and the results of L2 loss are better.

\begin{figure}[!t]
    \centering
    \includegraphics[width=\linewidth]{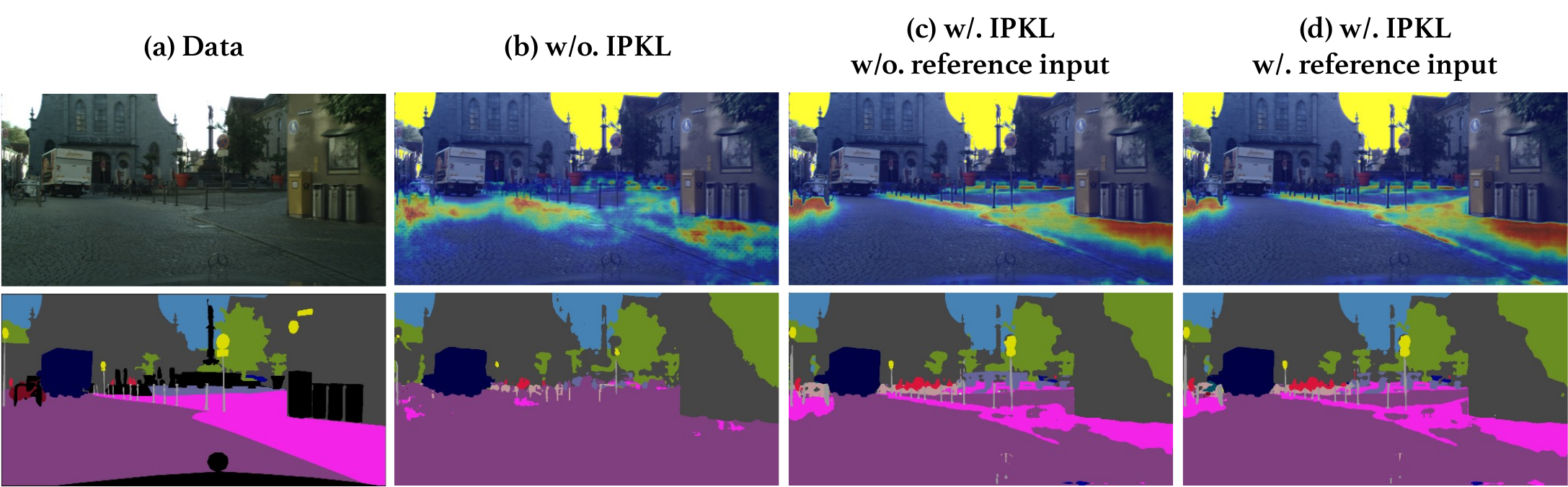} 
    \caption{Heatmap (on \textit{sidewalk}) and segmentation results comparision between (b) w/o. IPKL; (c) w/. IPKL and w/o. reference input; (d) w/. IPKL and w/. reference input.}
    \label{fig:ipkl study}
    \vspace{-0.2cm}
\end{figure}

As shown in Fig.~\ref{fig:ipkl study}, by introducing IPKL, the model could predict results (c) on the unseen data without a guided reference input similar to those (d) with a guided reference input.
Both of them are significantly superior to the results without IPKL (b).

\begin{table}[!t]
    \centering
    \caption{
        Ablation study on the proposed components DIFF and IPKL with different consistency losses. 
    }
    \label{tbl:ablation study}
    \resizebox{0.7\columnwidth}{!}{
    \begin{tabular}{ccc|c}
        \toprule
        \multirow{2}{*}{DIFF}  & \multirow{2}{*}{IPKL}    &  \multirow{2}{*}{$\mathcal{L}_\text{consis}$}  &  \multicolumn{1}{c}{mIoU (\%) on}  \\
        \cmidrule{4-4}
        &  &  & GTAV~\cite{richterPlayingDataGround2016} $\rightarrow$ CS~\cite{cordtsCityscapesDatasetSemantic} \\
        \midrule
            --       &  --  &   --       &   49.72 (baseline)        \\
        \checkmark   &  --  &   --       &   56.28 (\textcolor{red}{+6.56})         \\
        \midrule
        \checkmark   &  \checkmark  &    w/o. $\mathcal{L}_\text{consis}$       &   44.82 (\textcolor{green}{-4.90})          \\
        \checkmark   &  \checkmark  &    KL Loss        &   57.81 (\textcolor{red}{+8.09})          \\
        \textbf{\checkmark}   &  \checkmark  &    \textbf{L2 Loss}     &   \textbf{58.01 (\textcolor{red}{+8.29})}    \\
        \bottomrule
    \end{tabular}
    }
    \vspace{-0.2cm}
\end{table}

\vspace{-0.2cm}
\section{Conclusion}
This paper delves into the potential of representations from pre-trained diffusion models in the challenging context of domain generalization for semantic segmentation.
By introducing DIffusion Feature Fusion (DIFF) and implicit posterior knowledge learning (IPKL), the network could uniquely model cross-domain features by leveraging the rich prior knowledge of pre-trained diffusion models.
Extensive experiments on multiple settings demonstrated the superior performance of our method compared to the existing domain generalization semantic segmentation methods.

\clearpage
\bibliographystyle{IEEEtran}
\bibliography{ref}

\end{document}